\documentclass{article}%
\usepackage[T1]{fontenc}%
\usepackage[utf8]{inputenc}%
\usepackage{lmodern}%
\usepackage{textcomp}%
\usepackage{lastpage}%
\usepackage{array}
\usepackage{geometry}%
\usepackage[ruled]{algorithm2e}

\usepackage{multicol}%
\usepackage{graphicx}%
\usepackage{lipsum}%
\usepackage{caption}%
\usepackage{booktabs}%
\usepackage{ragged2e}%
\usepackage{breqn}  
\title{Feature selection or extraction decision process for clustering using PCA and FRSD}%
\author{Jean{-}Sébastien Dessureault, Daniel Massicotte}%
\date{\today}%
\begin{document}%
\normalsize%
\maketitle%
\justify%
\section*{ABSTRACT}%
\label{sec:ABSTRACT}%
This paper concerns the critical decision process of extracting or selecting the features before applying a clustering algorithm. It is not obvious to evaluate the importance of the features since the most popular methods to do it are usually made for a supervised learning technique process. A clustering algorithm is an unsupervised method. It means that there is no known output label to match the input data. This paper proposes a new method to choose the best dimensionality reduction method (selection or extraction) according to the data scientist's parameters, aiming to apply a clustering process at the end. It uses Feature Ranking Process Based on Silhouette Decomposition (FRSD) algorithm, a Principal Component Analysis (PCA) algorithm, and a K-Means algorithm along with its metric, the Silhouette Index (SI).  This paper presents 5 use cases based on a smart city dataset.  This research also aims to discuss the impacts, the advantages, and the disadvantages of each choice that can be made in this unsupervised learning process.
\newline

\noindent Keywords: Feature extraction, feature selection, FRSD algorithm, PCA algorithm, k-mean algorithm, silhouette index

\section{Introduction}%
\label{sec:Introduction}%

There are several combinations of algorithms that can be used to prepare the features. The choice of the right combination is not obvious, since each one has its pros and cons.  There are different schools of thought when it's time to reduce dimensionality, extract or select the features, or leave the totality of the features intact. This research proposes a new method with its own input, output, and metrics to make the right choice between feature extraction and feature selection. According to the input parameters, the right combination of algorithms is used to do the dimensionality reduction, followed by a clustering process. It is based on the machine learning methods FRSD, PCA, k-means combined to its SI metric.

This new method can be used in different contexts. In this paper, the features used to test this method were about smart cities. At the age of smart cities and intelligent urbanism, there is a need to analyze the data and to have an excellent knowledge of them \cite{noauthor_towards_nodate} \cite{kitchin_real-time_2014}. In this specific context, an important part of the challenge comes from the fact that the source of the data can be multiple. Data can come from different censuses, from local health organisms, from local dwelling organisms, from the economic sector, and so on. Some research like \cite{tayyebi_urban_2011} \cite{noauthor_modeling_2013} in the smart urbanism area already showed the importance of having a good understanding of the data. 

Principal Component Analysis (PCA) is a useful algorithm to extract features and reduce the dimensionality of a dataset \cite{noauthor_spectral_nodate}. It has been also used in a smart city context \cite{noauthor_sustainability_nodate}. It consists of linear transformations to convert a set of correlated variables into a set of linearly uncorrelated variables. Wong \cite{wong_developing_2016} showed that a PCA algorithm can help determine an indicator like a local economic development (LED) indicator. He defined a framework of 11 features, having first a total of 29 features. He used regression models to find relative strengths of the relationship between the LED indicator and performance variables.  Others like \cite{chen_city_2008} use a PCA algorithm combined with cluster analysis (CA) to study social-economic indexes (i.e., non-agriculture population; gross industry output value; business volume of post and telecommunications; local governments revenue, and others.) The analysis is applied to 17 counties and cities.  In this example, a PCA algorithm is used to retrieve the first and the second principal component (PC1 and PC2). According to PC1 and PC2, the CA classifies the cities into four classes of growth poles. Researches like \cite{noauthor_big_nodate}\cite{noauthor_lessons_2015} also use a PCA algorithm to extract features, in the field of big data and smart cities.  

As the reduction of dimensionality, clustering is an important part of the unsupervised learning process. In a smart urbanism context, it has been widely used over the years to regroup similar parts of a territory together. Different algorithms can be used, like k-means for standard crispy clustering \cite{8691036}, or c-mean for fuzzy clustering \cite{abed_identifying_2011}\cite{noauthor_modeling_2013}. There are several techniques to evaluate the consistency of the generated clusters \cite{desgraupes2013clustering}. One of them is the Silhouette Index (SI) \cite{Rousseeuw_2009} \cite{rousseeuw1987silhouettes}.
This SI metric for clustering has been used by \cite{dessureault_unsupervised_nodate} in a smart urbanism context. 

To evaluate the importance of the features in an unsupervised learning context, we have to first generate the label according to one criterion. In 2020,  \cite{yu_ensemble_2020} proposed the Feature Ranking Process Based on Silhouette Decomposition (FRSD) algorithm. It aims to solve this evaluation of features for clustering using a Silhouette Index (SI) metric.  It consists of generating SI for possible every combination of features, for each value of \textit{k} (the number of clusters) in a k-means clustering algorithm. 

The main contribution of this paper is to propose a new machine learning method to automatize the reduction of the dimensionality decision process (feature extraction or feature selection) of a clustering algorithm, according to the data scientist preferences. 

The dataset used in the research comes from the London Datastore. It defines a deprivation index of each ward of the London area.  In Great Britain, a ward is known to be a geolocalisation unit. 

The next sections of this paper are organized with the following structure: Section \ref{sec:Methodology} describes the proposed methodology. Section \ref{sec:Results} presents the results. Section \ref{sec:Discussions} discusses the results and their meaning and Section \ref{sec:Conclusion} concludes this research.

\section{Methodology}%
\label{sec:Methodology}%

\subsection{Selected features}%
\label{subsec:Sub_methodology_Selected features}%
As previously mentioned, the used dataset comes from the London Datastore. It is called "Indices of Deprivation from the Ministry of Housing, Communities \& Local Government (MHCLG)". There are 4766 records in this dataset. 8 features have been kept for this research (resume in Table \ref{table:table_features} \cite{leeser_english_2015}. There is also the ward code, which is the unique identification of a geographical sector in the United Kingdom. 
1. IMD score is the Index of Multiple Deprivation. It is a combined index of other features. 2. The income deprivation score aims to give the proportion of people in an area who are living on low incomes. 3. The employment deprivation score is a simple proportion of working-age people who are involuntarily out of work – including those unable to work due to incapacity or disability. 4. The health deprivation score takes into account a wide range of aspects, including premature death and mental health issues as well as measures of morbidity and disability. 5. The education, skills, and training deprivation score are formed from two subdomains combined with equal weights.  The first includes measures for children and young people, using achievement and participation data at various educational stages. The second subdomain is a measure for working-age adults.  6. The barriers to housing and service score have two equally weighted subdomains – geographical barriers and wider barriers to suitable housing (household overcrowding, homelessness, etc.). 7. The crime score uses data on 33 types of recorded crime under four broad categories – burglary, theft, criminal damage, and violence. 8. The living environment deprivation score. It includes housing issues in terms of the standard of housing as the "indoors" living environment (central heating, poor conditions, etc.). 

Table \ref{table:table_features} shows the list of all the features that are processed by the machine learning methods. \newline%

\begin{minipage}[htb]{1.0\columnwidth}
\centering
\captionof{table}{List of the features of the wards in the Greater London area}%
\begin{tabular}{|l|l|} \hline 
\label{table:table_features}%
\textbf{Rank} & \textbf{Features} \\
\hline
1.&IMD Score \\
2.&Income Score\\
3.&Employment Score\\
4.&Health Score\\
5.&Education Score\\
6.&Barriers Score\\
7.&Crime Score\\
8.&Living Score\\
\hline
\end{tabular}
\end{minipage}\newline

There are 630 wards in the Greater London area. Section \ref{subsec:CityOfLondon} gives more explanation about the city of London and the ward system. 

To visually represent the features, a radar graphic of a typical ward (00ADGM) is shown in Fig. \ref{fig:radar1}. It shows the 8 features values on 8 different axes. All the values are normalized using a MinMax function to fit the graph scale from 0 to 1. The higher the values, the better the score is for each feature. 

\noindent \includegraphics[width=10cm]{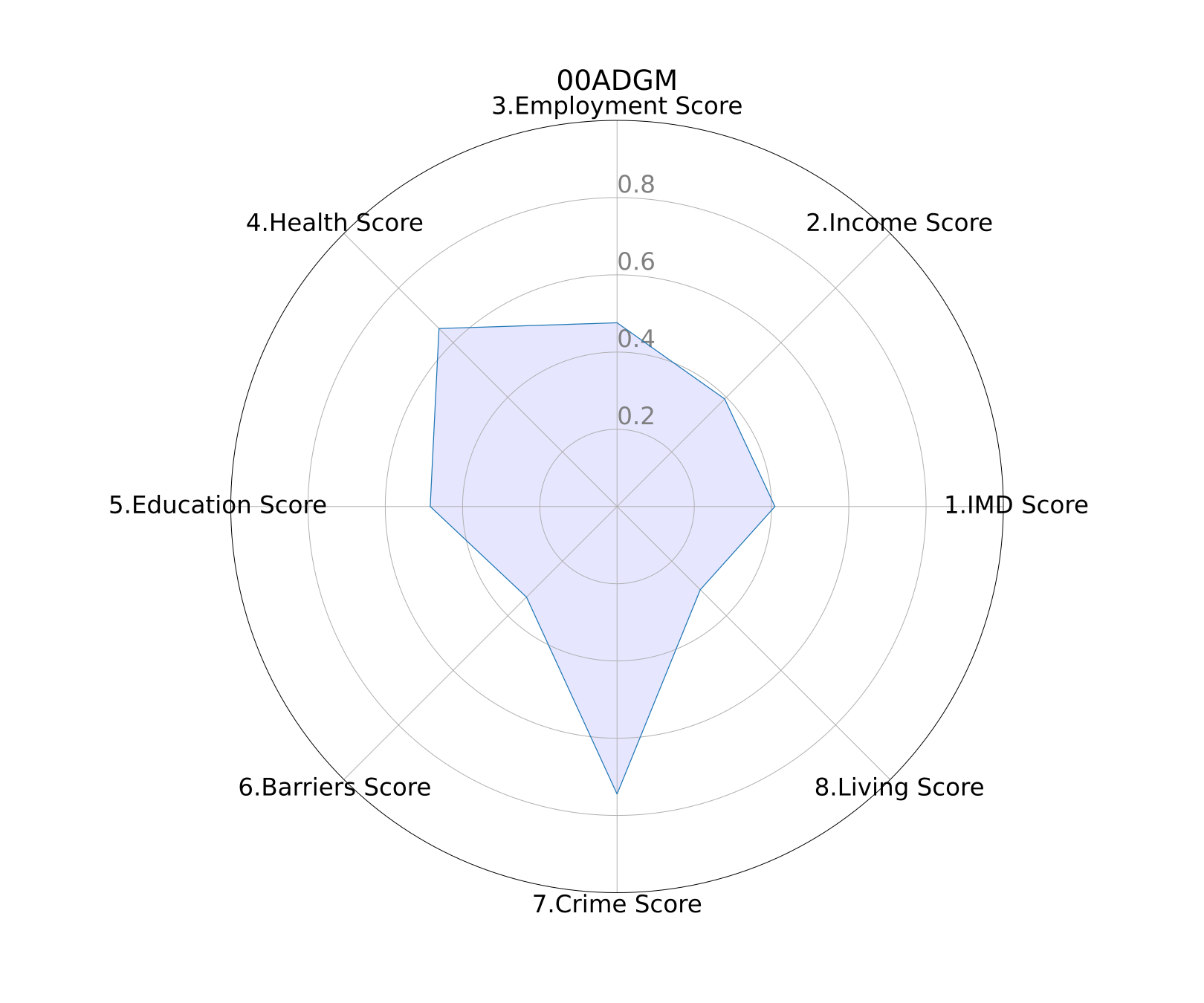}
\centering
\captionof{figure}{Radar graphic representing every MinMax normalized feature  \label{fig:radar1}\newline}%
\justifying

\subsection{Proposed model design}%
\label{subsec:Sub_methodology_model_design}%
This new model includes several parts that lead to a complete parametrized process. Fig. \ref{fig:architecture} presents the architecture of the complete methodology. The first part is a 3 steps method to evaluate the importance of each feature in a clustering process using an FRSD algorithm. Step 1 is a loop that generates a SI from every feature combination. Step 2 Aggregates the results. Step 3 Normalizes the importance of the features using a \textit{MinMax} algorithm. 
The second part evaluates the feature importance according to a PCA algorithm.  It converts the features into principal components (PCs) allowing the evaluation of the explained variance contribution of each feature.  
After having calculated the feature importance in an unsupervised machine learning context using both FRSD and PCA, the method has to select between a feature selection (FS) or a feature extraction (FE) according to some parameters defined by the user. It calculates two scores. Both scores are calculated according to the input parameters of the algorithm. Those scores and their equations will be defined in section \ref{subsec:Sub_choice_FE_FS}.  The user preferences parameters allow the algorithm to orient the results toward 1. Interpretability or 2. Integrity.  
Having those two scores, the process has to choose between the best option, knowing the data and the user's preferences for interpretability and integrity. A higher score in interpretability leads to a feature selection.  A higher score in integrity leads to a feature extraction using PCA.  
After having made a reduction of dimensionality (using a FS or a FE),  clustering is applied using a K-Means algorithm, returning the output to the user.  The method also includes a normalization (using a MinMax algorithm) of the output and the production of stacked radar graphics (a stacking of several graphs presented in Fig. \ref{fig:radar1}) to illustrate better the result of the clustering process.   

The features used to test this model are the ones presented in Table \ref{table:table_features}. The following section describes each part of the process and its machine learning algorithms. \newline

\noindent \includegraphics[width=16cm]{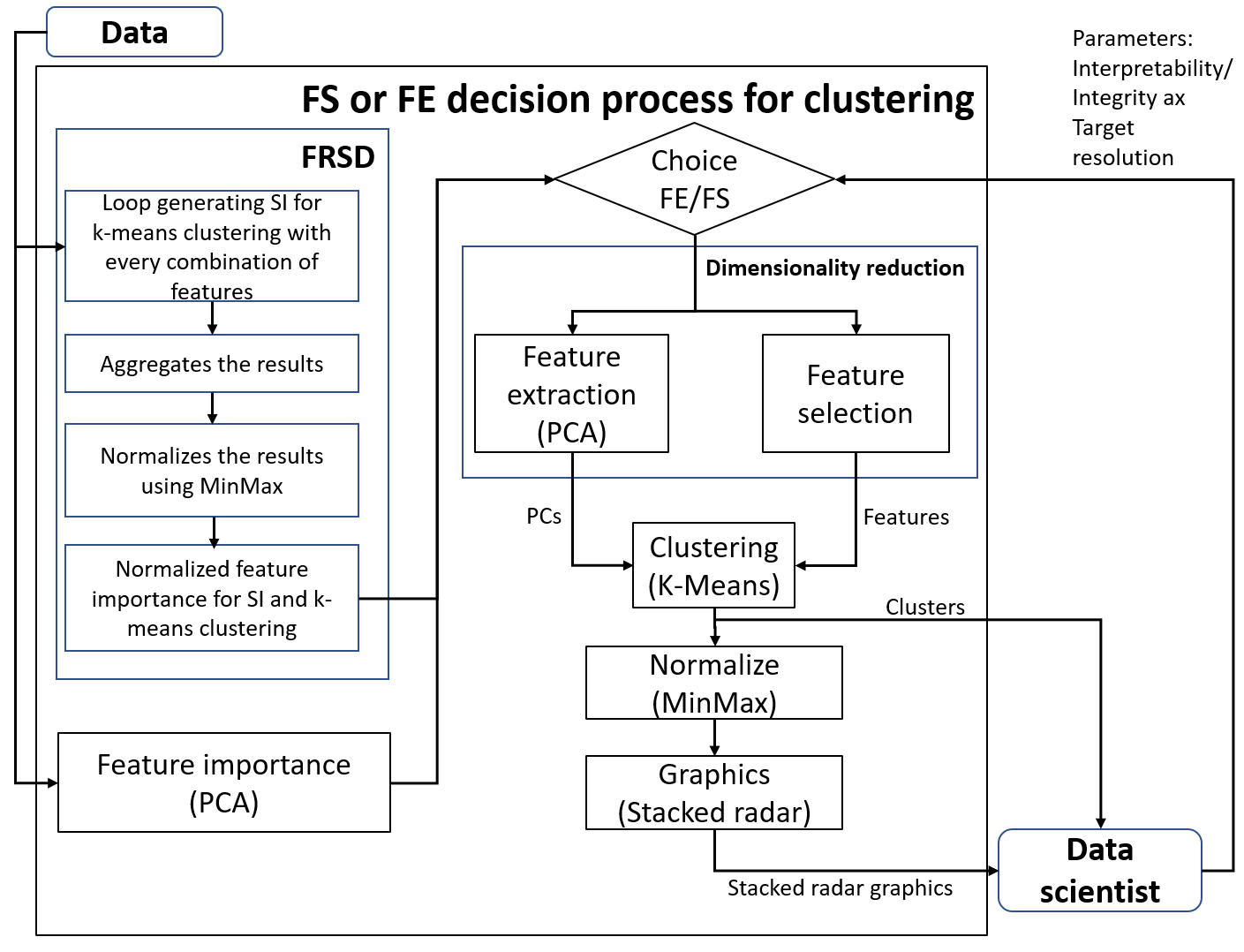}
\captionof{figure}{Architecture of the methodology \newline \label{fig:architecture}}%

\subsection{Evaluation of features using FRSD}%
\label{subsec:Sub_methodology_FRSD}%
It is more complicated to analyze the importance of the features in an unsupervised context than in a supervised context, which not only needs the data but also needs the label of each data. In an unsupervised context, the problem must be considered differently because of two main purposes:
1. The feature number can be variable. For the city of London, the number of features can be between 2 and 8 features. 2. In the case of a clustering algorithm like k-means, there is no label since it is an unsupervised technique.\newline
Since there is no labeled data, it must be generated. The SI metric is used to evaluate the consistency of the clustering. In this method, it is also used to label the unlabeled features. 

The FRSD algorithm calculates the importance of each feature by decomposing the average silhouette widths of the random subspaces. The goal is to solve this evaluation of features importance for clustering using a SI metric. The pseudo-code is presented in Algorithm  \ref{algo:algoFRSD}.

\begin{algorithm}[H]
\SetAlgoLined
 initialization\;
 \For{nb_cluster in range self.k_min to self.k_max}{
 \For{subset in all possible subset of length 2 to maximum length}{
 SI = k-means(subset, nb_cluster)\;
 Add_results(subset, SI) \;
  }
 }
 initialize feature synthesis \;
 \For{element in results}{
 \For{feature in features_list}{
 \If{feature exists in element}{
 add the result value (SI) to the feature synthesis\;
 } 
 }
 }
 normalize feature synthesis using MinMax\;
 \caption{FRSD algorithm}
 \label{algo:algoFRSD}
\end{algorithm}

The complete FRSD process is declined in three steps \cite{yu_ensemble_2020}.\newline

SI generation:\newline
This step aims to generate de SI values for each combination of features and for each evaluated cluster number \textit{k} in a determined range. To determine it, a loop from \textit{a minimum number of possible clusters} to \textit{a maximum number of possible clusters} must be executed.  A range of 3 to 15 has been used in this method to generated the SI. Inside this loop, a second loop generates a SI from every combination of the features. In the case of the city of London, a minimum of 2 and a maximum of 8 features are needed.  
In this particular case, it means 248 results (possible combinations between 2 and 8 features), \textit{i} times from i in range $k_{min}$ and $k_{max}$. Since $k_{max}$ and $k_{min}$ = 12, a total of 148 * 12 = 1776 SI indexes will be created. Table \ref{table:TblSilhouette1} shows a partial example of the output generated for \textit{k}=10.\newline

\begin{minipage}[htb]{1.0\columnwidth}
\centering
\captionof{table}{Generation of Silhouette index for each combination of the features.}%
\begin{tabular}{|l|l|} \hline 
\label{table:TblSilhouette1}%
\textbf{Partial list of features} & \textbf{SI} \\ \hline 
1,2&0.6084\\
1,3&0.6085\\
1,4&0.6066\\
1,5&0.5037\\
1,6&0.4658\\
1,7&0.6069\\
1,8&0.4783\\
2,3&0.5803\\
...&...\\
1,3,4,5,6,7,8&0.2833\\
2,3,4,5,6,7,8&0.2837\\
1,2,3,4,5,6,7,8&0.2873 \\ \hline 
\end{tabular}
\end{minipage}\newline

SI aggregation:\newline 
The algorithm aggregates the total of the SI for every feature.  It sums the SI value if feature \textit{n} is in the feature's list used to compute this SI. The result is a vector the same size as the number of features. For each index, the sum of the SI for this feature index. This sum is divided by the sum of all the features.\newline

Weighting feature:\newline
The final ratio representing the feature importance is available in the final vector. Applying a \textit{MinMax} function in the final vector will help to discriminate the values and improve the presentation. Equation (\ref{eq:MinMax}) shows the MinMax normalization formula.  It simply normalizes a number to get a 0 to 1 range, associating the smallest value to 0 and the highest to 1. In (\ref{eq:MinMax}), \textit{x} is the input value to normalize.  

\begin{equation}%
z = \frac{x - min(x)}{max(x) - min(x)}%
 \label{eq:MinMax}%
\end{equation}%

The result is the importance of each feature in the clustering process. It can be summarized in \ref{eq:FRSD1} and \ref{eq:FRSD2}. 

\begin{equation}%
R_{sub, SI} = \sum _{k=k _{ min}}^{k _{max}}\sum _{sub=0}^{max(sub)} kmeans(sub, k) 
 \label{eq:FRSD1}%
\end{equation}%
Where \textit{R} is the resulting matrix, \textit{sub} is the possibles subsets of features, \textit{SI} is the Silhouette index, \textit{k} is the number of clusters.  

\begin{equation}%
NFI = \sum _{i=0}^{max(R)}\sum _{feat=0}^{feat.in R} SI 
 \label{eq:FRSD2}%
\end{equation}%

Where \textit{NFI} is the normalized feature importance, \textit{SI} is the Silhouette index, \textit{R} is the resulting matrix of \ref{eq:FRSD1}.  

The final result, a list of features and their importance will be used in the decision process documented in section \ref{subsec:Sub_choice_FE_FS}.

\subsection{Evaluation of features using PCA}%
\label{subsec:Sub_methodology_PCA_evaluation}%
This part simply runs a PCA algorithm on all the features.  The number of PCs specified is the same as the number of features in the input. Among other results, it gives a result in terms of explained variance for each feature.  Aggregating this variable for each feature of each PCs, and dividing par the total amount of explained variance, gives the importance of each feature in a data extraction process. 

This list of features and their importance will be important in the decision process documented in section \ref{subsec:Sub_choice_FE_FS}.

\subsection{Choice between feature extraction and feature selection}%
\label{subsec:Sub_choice_FE_FS}%
This part of the whole algorithm needs some parameters defined by the user according to his features requirements. Before defining the parameters, let's define an important axis about dimensionality reduction.  This is the "Interpretability/integrity" axis. When choosing the method of dimensionality reduction, we have to choose between optimizing the interpretability of the features, or the integrity of the features. To optimize the interpretability of the features, a feature selection method must be used. Using this method, every feature will keep its name and significance. Although, some features are completely dropped, causing a reduction in the resolution of the data. On the opposite, to optimize the integrity of the features, a feature extraction method (like PCA) must be used. Using this method, every feature is used to generate a new set of normalized data. Since every feature is used in this process, the integrity of the data is better than with a feature selection that completely drops some of them.  The counterpart of this method is that the feature's names are lost, being replaced by "principal components". Consequently, there is a loss of features interpretability. 

In this method, two parameters are defining this "interpretability/integrity" axis: 1. interpretability_oriented and  2. integrity_oriented. Both domains are a normalized number between 0 and 1, representing a percentage of importance. The sum of those numbers must equal 1. It simply describes how important this is. For instance, a value of 0.1 means "not very important" and a value of 0.9 means "very important". 

Another parameter is the target resolution (target_resolution). This one is used by the algorithm to select the right amount of features in the reduction of the dimensionality process.  Just enough features are kept to reach this resolution target. A high value means more features and a low value means fewer features. 

There are also two more important parameters: the minimum and the maximum k parameter of the K-Means algorithm (\textit{k_min} and \textit{k_max}). In other words, it defines the domain of the possible number of clusters.

Now that we have defined every important parameter, let's define this decision part of the algorithm itself. First, the algorithm selects only the best features that reach the minimum resolution, based on the FRSD process.  Then, the algorithm tries every possible value of k (number of clusters) between the range \textit{k_min} and \textit{k_max}.  It keeps the value of \textit{k} resulting in the higher value of the SI. Using the interpretability_oriented parameter and the best-found value of SI (the "best SI" variable) in the clustering process, it computes the "interpretability score" as defined in (\ref{eq:interpretability_score})

\begin{equation}%
interpret.score = interpret.oriented * best SI 
 \label{eq:interpretability_score}%
\end{equation}%

The next part consists of finding the "integrity score". It uses the PCA feature extraction algorithm. The algorithm uses only the required number of features to reach a minimum resolution, according to the PCA importance feature process. Then, it loops on every possible value of \textit{k} in the \textit{k_min} and \textit{k_max} range. Containing the best result of consistency in the clustering process, the best value of SI ("best SI" variable) is kept and the "integrity score" is computed as defined in (\ref{eq:integrity_score})

\begin{equation}%
integrity.score = integrity.oriented * best SI 
 \label{eq:integrity_score}%
\end{equation}%

The algorithm compares the two scores and selects the one having the greater value as an orientation for the dimensionality reduction. These are the possible cases: 1. The interpretability score is higher than the integrity score. A feature selection is done in this case. This process consists in keeping just enough features to reach the data resolution parameter. The others are dropped and lost. 2. The integrity score is higher than the interpretability score. In that case, a feature extraction must be done. This process is more complex than the simple feature selection. This process is explained in \ref{subsec:Sub_methodology_PCA}. 

The results of the decision are ultimately displayed to the user to justify the algorithm choice. Those values are 1. Normalized synthesis of features after the FRSD process 2. Normalized synthesis of features after the PCA process 3. Best SI for Feature selection 4. Best SI for feature extraction 5. Interpretability score 6. Integrity score 7. Chosen method (selection of extraction) 8. Number of selected features to obtain the target resolution (if feature selection is used) 9. Number of principal components to obtain the target resolution (if feature extraction is used) 10. Best number of clusters (k). \newline

\subsection{Dimensionality reduction using PCA}%
\label{subsec:Sub_methodology_PCA}%
A PCA algorithm aims to reduce the dimensionality of the dataset by extracting some features. It creates a new dataset having an equal number or less dimension than the original dataset. The newly created features are named "principal components" (PCs). The first principal component (PC1) has the highest possible variance compared to the other principal components.  The second principal component (PC2) has the second-highest possible variance, and so on. A PCA algorithm uses the concept of Eigen Vector and Eigen
Value. It compares every possible combination of 2 features. For every pair of features, it calculates the direction of the data distribution (the Eigen Vector) and the magnitude of this vector (the Eigen Value). A projection of the data is made using the axis of the strongest Eigen Value. At the end of this process, a descending ordered list of features is produced, based on the Eigen Value criterion. The PCA algorithm extracts the most n significant components, where n is a received parameter.

\subsection{Clustering with k-means}%
\label{subsec:Sub_kmeans}%
The goal of this process is to create some clusters after having reduced the dimensionality, based on the data and some parameters. After having reduced the dimensionality using a feature selection or a feature extraction, a clustering algorithm must be used. 
To create some clusters out of the data, it is necessary to use an unsupervised learning technique since there is no label for each input data. This algorithm will assign to each ward a reference cluster, according to the similarity level of their features. 

\noindent Equation (\ref{eq:kmeans}) defined the \textit{k}-means clustering equation where \textit{J} is a clustering function, \textit{k} is the number of clusters, \textit{n} is the number of features, $x_{i}^{(j)}$ is the input (feature \textit{i} in cluster \textit{j}) and $c_{j}$ is the centroid for cluster \textit{j}. Centroids are obtained by randomly trying some values and selecting the best according to the returned inertia value. This inertia value is the basic non-normalized metric to evaluate the cluster consistency.  

\begin{equation} \label{eq:kmeans} 
J=\sum _{j=1}^{k}\sum _{i=1}^{n}\left\| x_{i}^{(j)} -c_{j} \right\| ^{2}    
\end{equation} 

Exists several metrics to measure a clustering performance. Each metric is not necessarily compatible with every clustering algorithm.  Since the k-means algorithm is used, the clustering performance has been measured by the SI metrics. This metric is documented in \cite{Rousseeuw_2009} and \cite{gueorguieva_mmfcm_2017}. The SI is defined by 3 equations. First, the distance between each point and the center of its cluster is defined by (\ref{eq:Silhouette1}). The distance between the center of each cluster is shown in (\ref{eq:Silhouette2}). Finally, (\ref{eq:Silhouette3}) uses the result of (\ref{eq:Silhouette1}) and (\ref{eq:Silhouette2}) to calculate the final SI score that indicates the quality (the consistency) of the clustering process. SI ranges from -1 to +1. Values from -1 to 0 indicates bad classification. SI values from 0 to 1 indicate the points associated with a good cluster. The higher the value, the better is the cluster consistency \cite{Rousseeuw_2009}. 

\begin{equation} \label{eq:Silhouette1} 
a(i)=\frac{1}{\left|c_{i} -1\right|} \sum _{j\in c_{i} i\ne j}d(i,j)  
\end{equation} 
\begin{equation} \label{eq:Silhouette2} 
b(i)={\mathop{\min }\limits_{k\ne i}} \frac{1}{\left|c_{k} \right|} \sum _{j\in c_{k} }d(i,j)  
\end{equation} 
\begin{equation} \label{eq:Silhouette3} 
s(i)=\frac{b(i)-a(i)}{\max \left(a(i),b(i)\right)} ,\; \; {\rm if}\; \left|C_{i} \right|>1 
\end{equation} 

As shown in Fig \ref{fig:architecture}, there are two processes (1. Normalize and 2. Graphics) added after the clustering. Both are required for the visualization of the clustering process. A good way to visually represent the consistency of the clusters is by stacking the radar graphics representing their features. A MinMax algorithm (\ref{eq:MinMax}) must precede this type of graphic.  Using this type of graphic, the different profiles of the clusters are notable. This representation will be useful in \ref{subsec:SubResultsScenarios}.  

\section{Results}%
\label{sec:Results}%

\subsection{City of London}%
\label{subsec:CityOfLondon}%
The city of London is the capital of England and the United Kingdom.  It is also the largest city in the country. It stands on the River Thames and it exists since the Roman era. In the London metropolitan area, there were 14,040,163 inhabitants in 2016. The United Kingdom territory is divided into wards and electoral divisions. The ward is the primary unit of English electoral geography for civil parishes and borough and district councils. Each ward/division has an average electorate of about 5,500 people, but ward-population counts can vary substantially. At the end of 2014, there were 9,456 electoral wards/divisions in the United Kingdom \cite{council_ward_2013}.  \newline

Fig. \ref{fig:GreaterLondon} displays a map of the wards in the Greater London area. \newline

\centering
\noindent \includegraphics[width=6.5cm]{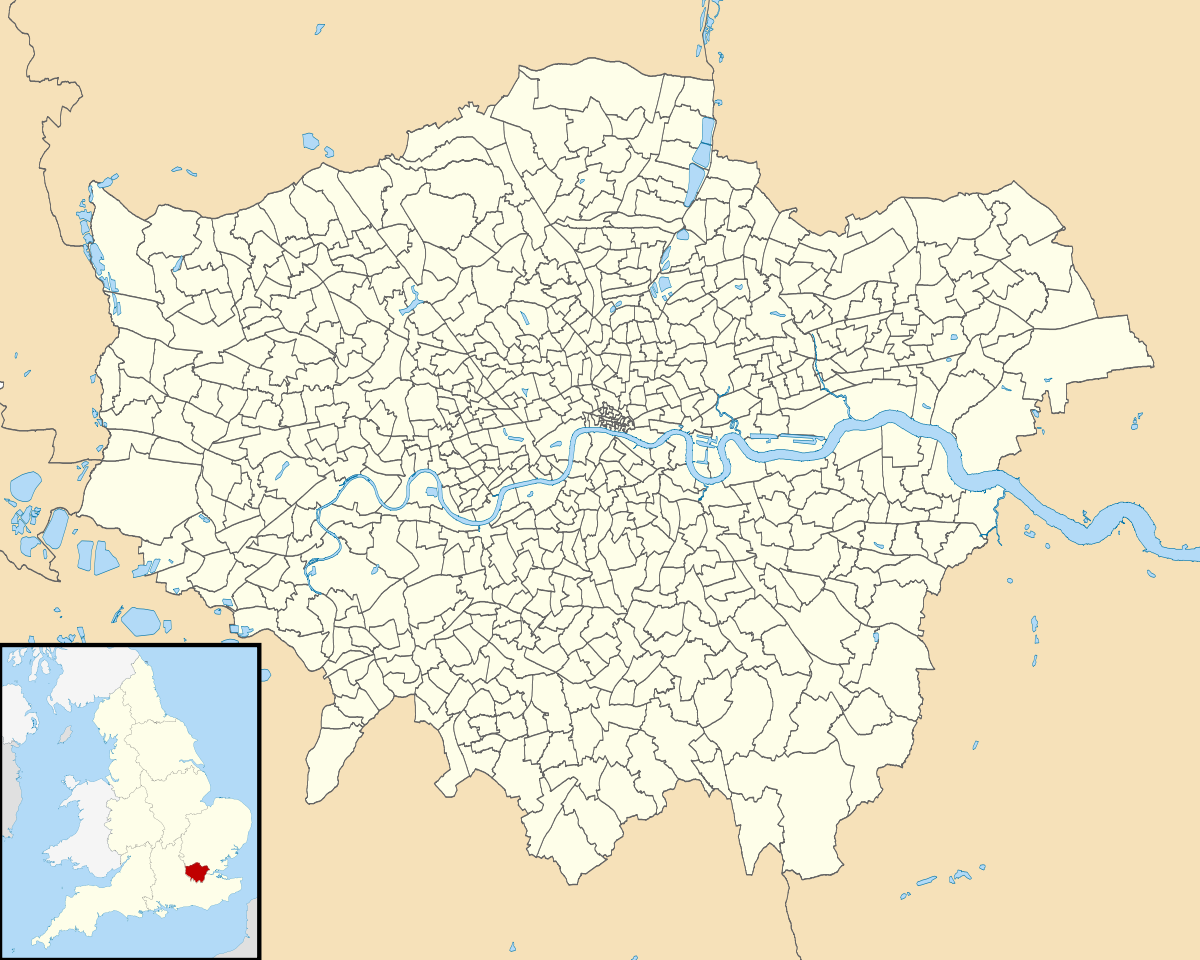}
\captionof{figure}{Map of the ward divisions of the Greater London area  \cite{noauthor_filegreater_nodate} \label{fig:GreaterLondon}\newline}%
\justifying

Sections \ref{subsec:SubResultsFeatureImportance_FRSD}, \ref{subsec:SubResultsFeatureImportancePCA}, and \ref{subsec:SubResultsScenarios} present the results of the methodology presented in section \ref{sec:Methodology} applied to the wards of the Greater London area. \newline

\subsection{Feature importance according to FRSD}%
\label{subsec:SubResultsFeatureImportance_FRSD}%
No matter what the parameters are, this part will be used to calculate the importance of the features for selection regarding the consistency of the clustering. The result of this part is useful in the dimensionality reduction process, to decide between feature selection and feature extraction. As defined in the methodology, an unsupervised approach is different from a supervised approach in the evaluation of the importance of the features. To find it on unlabeled data, we must find a way to generate labels. One state-of-the-art way to do it is to use a FRSD algorithm. The SI metric has been generated for each data, so it became possible to evaluate the features in a supervised learning way.

Table \ref{table:Feat_importance_FRSD} shows the list of all the features in importance order according to the FRSD evaluation. \newline

\begin{minipage}[htb]{1.0\columnwidth}
\centering
\captionof{table}{Feature importance according to FRSD}%
\label{table:Feat_importance_FRSD}
\begin{tabular}{|l|l|r|}
\hline 
\textbf{\#} & \textbf{Features} & \textbf{Norm. weights} \\ 
\hline  
  1 &  3.Employment Score &    0.1319 \\
  2 &      2.Income Score &    0.1315 \\
  3 &       7.Crime Score &    0.1298 \\
  4 &      4.Health Score &    0.1294 \\
  5 &         1.IMD Score &    0.1217 \\
  6 &   5.Education Score &    0.1205 \\
  7 &      8.Living Score &    0.1178 \\
  8 &    6.Barriers Score &    0.1172 \\
\hline 
\end{tabular}
\end{minipage}\newline

\subsection{Feature importance according to PCA}%
\label{subsec:SubResultsFeatureImportancePCA}%
This part is used independently of the parameters, to calculate the feature importance in a feature extraction process. This is important to choose between a feature selection and a feature extraction. The PCA algorithm, as documented in the methodology, has been applied to the 8 features in the London dataset. 

Table \ref{table:Feat_importance_PCA} shows the list of all the features in importance order, according to the PCA evaluation. \newline

\begin{minipage}[htb]{1.0\columnwidth}
\centering
\captionof{table}{Feature importance according to the PCA algorithm}%
\label{table:Feat_importance_PCA}
\begin{tabular}{|l|l|r|}
\hline 
\textbf{\#} & \textbf{Features} & \textbf{Norm. weights} \\  
\hline 
  1 &  PC1 &   0.1366 \\
  2 &  PC2 &   0.1365 \\
  3 &  PC3 &   0.1357 \\
  4 &  PC4 &   0.1329 \\
  5 &  PC5 &   0.1286 \\
  6 &  PC6 &   0.1222 \\
  7 &  PC7 &   0.1139 \\
  8 &  PC8 &   0.0931 \\
\hline 
\end{tabular}
\end{minipage}\newline

\subsection{Scenarios using different parameters}%
\label{subsec:SubResultsScenarios}%

This section describes five scenarios using different parameters values to test the new method. The common parameters for all scenarios are \textit{k_min} = 3 and \textit{k_max} = 10. It is the same for all the scenarios since it is useful to specify the number of possible clusters, but useless in the decision taken by the algorithm. The following scenarios show how the parameters affect the decision taken by the algorithm. \newline

\noindent \textbf{Case 1: Interpretability oriented and high resolution of data} \newline
For this first scenario, let's suppose that it is more important to keep the feature's names (interpretability) than it is to optimize the feature integrity. Also, let's suppose that a good feature resolution is needed.  The value of interpretability_oriented = 0.9, integrity_oriented = 0.1 and target_resolution = 85\% has been used. Table \ref{table:Scenario1} shows the results using this configuration. \newline

\begin{minipage}[htb]{1.0\columnwidth}
\centering
\captionof{table}{Algorithm results using interpretability_oriented = 0.9, integrity_oriented = 0.1 and target_resolution = 85\%}%
\label{table:Scenario1}
\begin{tabular}{|l|r|}
\hline 
\textbf{Metrics} & \textbf{Values} \\ 
\hline 
  Best FS silhouette index & 0.3905  \\
  Best FE silhouette index & 0.3530  \\
  Interpretability score 	  & 0.3514  \\
  Integrity score 		  & 0.0353  \\
  Chosen method 		  & SELECTION  \\
  Number of selected features &  \\  
  to obtain target resolution & 7  \\
  Resolution 			  & 88.3\%  \\
  Best number of clusters (k) & 3  \\
\hline 
\end{tabular}
\end{minipage}\newline

In this table, we can see that the value of the best feature selection (FS) silhouette index (0.3905) is greater than the value of the best feature extraction (FE) silhouette index (0.3530). 
Note that the consistency of the clustering process has nothing to do with the resolution of data. Often the more consistency comes with less dimension. It can be very hard to have a good consistency with a high number of features. That is why using this method, orienting a process toward integrity (by using feature extraction instead of feature selection) does not mean a better consistency while clustering. Often, lowering the resolution gives in a SI showing a better consistency.  

In this scenario, the parameter interpretability_oriented (0.9) has a higher value than the integrity_oriented value (0.1). When (\ref{eq:interpretability_score}) and (\ref{eq:integrity_score}) are applied, the interpretability score (0.3514) is higher than the integrity score (0.0353). To reach 85\% of the resolution, we must use the best 7 features.  Those features have a resolution of 88.3\%. Doing the clustering process, the optimal number of clusters is 3. Fig. \ref{fig:Scenario1_Silhouette} shows the distribution of each element according to its silhouette index.  

\centering
\noindent \includegraphics[width=10cm]{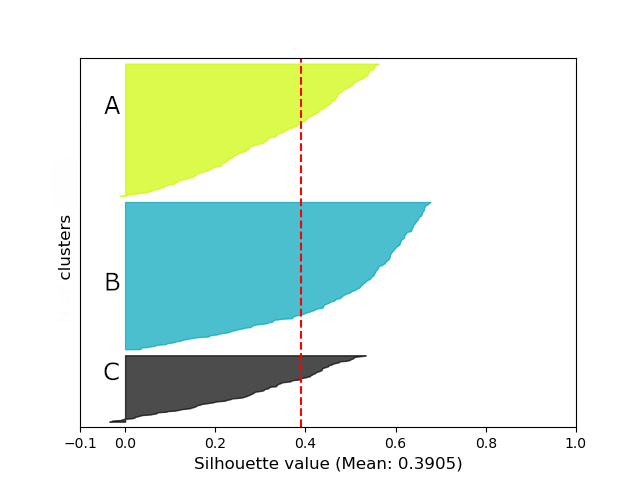}%
\captionof{figure}{Silhouette graphic for case 1 showing the consistency for clusters A, B, and C.\newline \label{fig:Scenario1_Silhouette}}
\justifying

This figure shows the 3 different clusters, in 3 different colors. The larger the horizontal bar is, the more the cluster contains data. The longer the bar is, the more consistent the data is according to its cluster. This graphic shows very few misplaced values (negative values). It also shows an average of 0.3905 (red dotted line).

Fig. \ref{fig:Scenario1_radar_multi_A} shows the representation of cluster A using a stacked radar graphic.  It is easy to visualize the consistency of the normalized value. It shows that the feature's names have been kept.  It is the most important criterion (the interpretability) for this case since the parameter interpretability_oriented is equal to 0.9. 

\centering
\noindent \includegraphics[width=10cm]{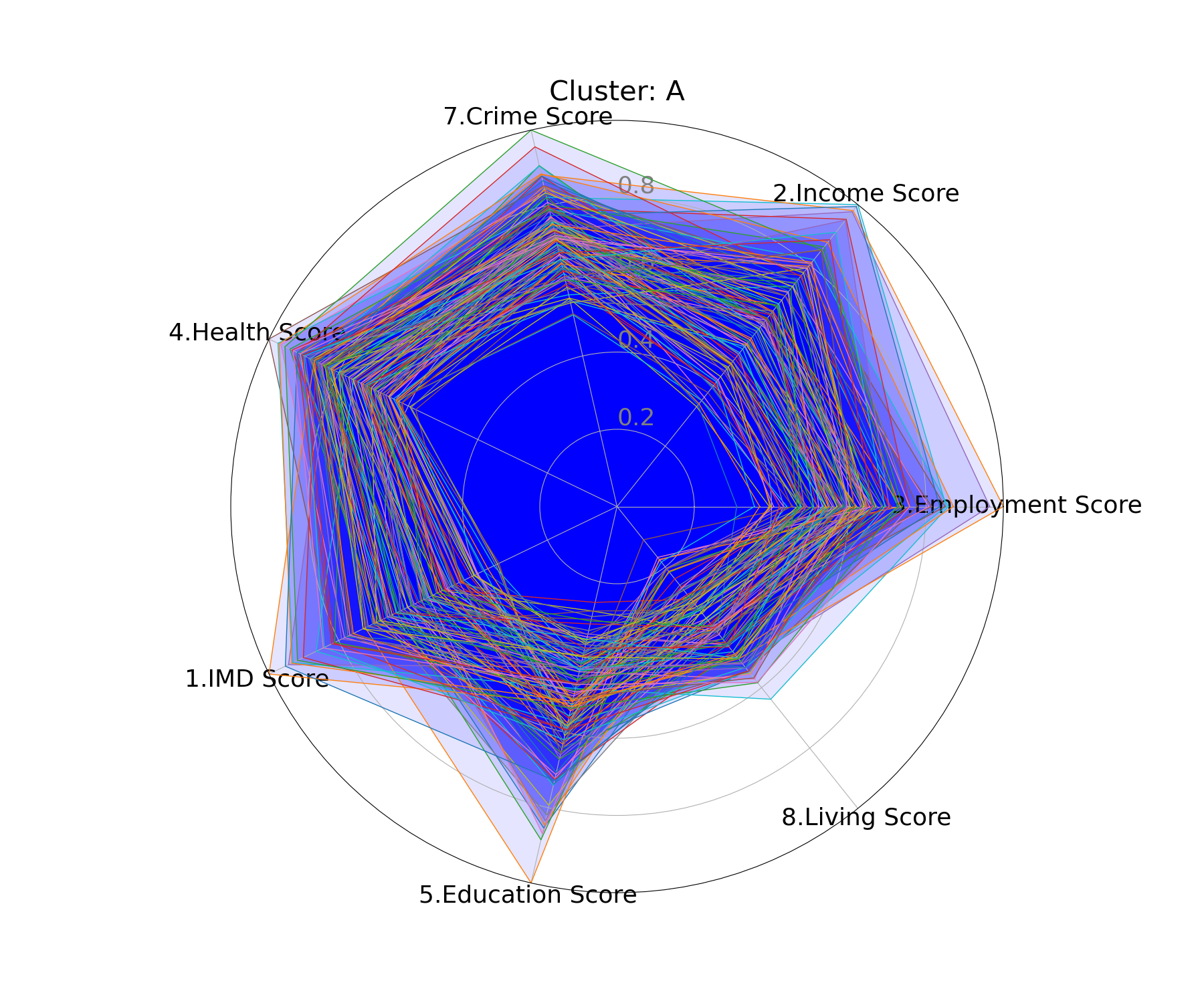}%
\captionof{figure}{Stacked radar graphics showing cluster A of normalized values for case 1\newline \label{fig:Scenario1_radar_multi_A}}
\justifying

\noindent \textbf{Case 2: Integrity oriented and high resolution of data} \newline
For this second scenario, integrity is more important than keeping the signification of the features (interpretability).  A good feature resolution is also needed.  The value of interpretability_oriented = 0.1, integrity_oriented = 0.9 and target_resolution = 85\% has been used.  Using this configuration, Table \ref{table:Scenario2} shows the results. \newline

\begin{minipage}[htb]{1.0\columnwidth}
\centering
\captionof{table}{Algorithm results using interpretability_oriented = 0.1, integrity_oriented = 0.9 and target_resolution = 85\%}%
\label{table:Scenario2}
\begin{tabular}{|l|r|}
\hline 
\textbf{Metrics} & \textbf{Values} \\ 
\hline 
  Best FS silhouette index & 0.3905  \\
  Best FE silhouette index & 0.3530  \\
  Interpretability score   & 0.0390  \\
  Integrity score 		  & 0.3177  \\
  Chosen method 		  & EXTRACTION  \\  
  Number of PCs 		  & 7  \\
  Resolution 			  & 90.7\%  \\
  Best number of clusters (k) & 3  \\
\hline 
\end{tabular}
\end{minipage}\newline

We can observe that the value of the best FE silhouette index (0.3530) is lower than the value of the best FS silhouette index (0.3905). Having an integrity parameter with a high value (0.9), the integrity score (0.3177) is higher than the interpretability score (0.0390). The feature extraction strategy is selected.  7 PCs are required to reach 85\% of resolution.  The optimal number of clusters is 3. Fig. \ref{fig:Scenario2_Silhouette} shows the distribution of the SI for this case.  

\centering
\noindent \includegraphics[width=10cm]{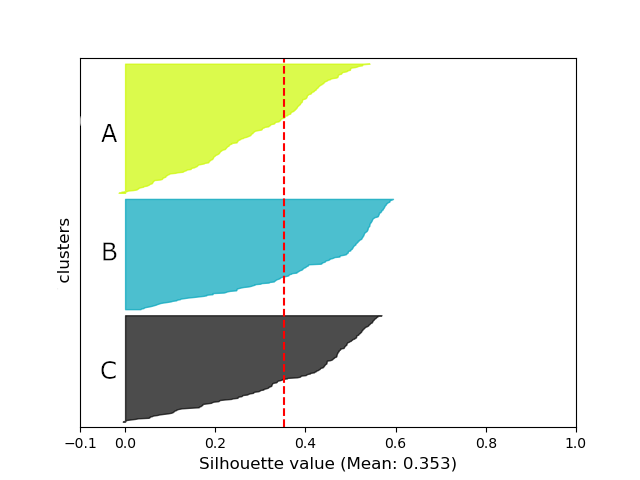}%
\captionof{figure}{Silhouette graphic for case 2 showing the consistency for clusters A, B, and C.\newline \label{fig:Scenario2_Silhouette}}
\justifying

This figure displays the 3 clusters. There are a few misplaced values (between -1 and 0). The average of the SI is 0.353. 
Even if the consistency of the clustering is lower, the integrity of the data is better since every feature has been used to downsize to the 7 PCs. The loss is at its minimum. Remind that it is often when reducing the dimensionality that the SI gets lower. For instance, having only 2 PCs or features tends to give the best SI results. 

Table \ref{fig:Scenario2_radar_multi_A} shows cluster A using a stacked radar graphic. 
  
\centering
\noindent \includegraphics[width=10cm]{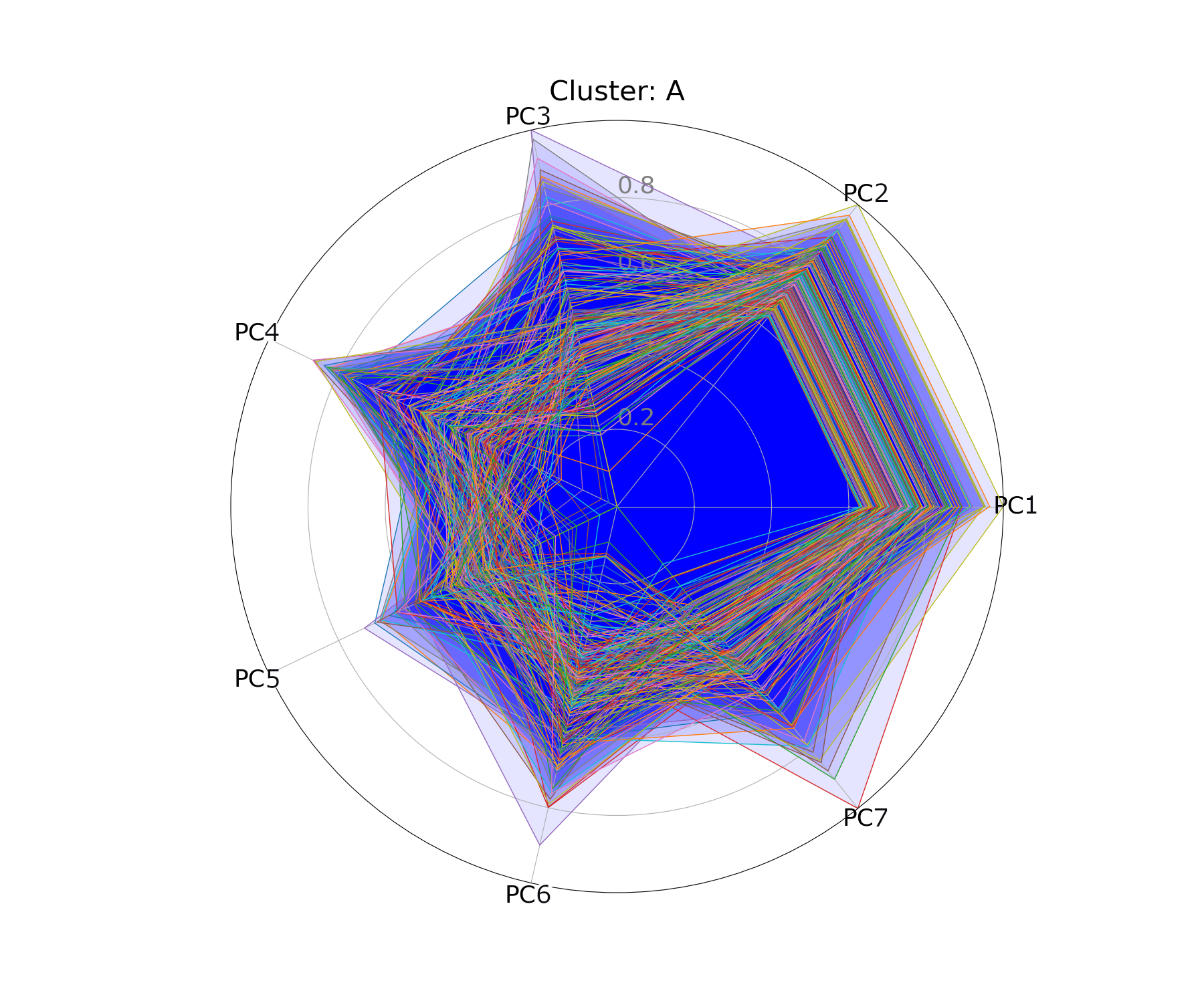}%
\captionof{figure}{Stacked radar graphics showing cluster A of normalized values for case 2\newline \label{fig:Scenario2_radar_multi_A}}
\justifying

It is important to keep in mind that when a feature extraction is made, all the feature's labels are lost.  In this particular case, for instance, it becomes impossible to refer to feature 5 "Education Score", since this value, like all others, has been extracted to generate the new features called "principal components" (PCs). Original features can no longer be addressed. It can be an important drawback, according to what has to be done next. For instance, if a clustering process is made (like in Fig. \ref{fig:Scenario2_radar_multi_A}), the clustering graphs would be represented having "PC1", "PC2", "PC3", and so on, on its axis. Having fewer dimensions is an advantage.  Losing the identity of the features is a disadvantage. That is the opposite of "interpretability".\newline

\noindent \textbf{Case 3: Equally integrity and interpretability oriented and high resolution of data} 
For this third scenario, we suppose that it is equally important to keep the features signification than it is to optimize the integrity of the features. We suppose that a good feature resolution is also needed. The value of interpretability_oriented is 0.5, integrity_oriented is 0.5 and target_resolution is 85\%. Table \ref{table:Scenario3} shows the results for this configuration. \newline

\begin{minipage}[htb]{1.0\columnwidth}
\centering
\captionof{table}{Algorithm results using interpretability_oriented = 0.5, integrity_oriented = 0.5 and target_resolution = 85\%}%
\label{table:Scenario3}
\begin{tabular}{|l|r|}
\hline 
\textbf{Metrics} & \textbf{Values} \\ 
\hline 
  Best FS silhouette index & 0.3905  \\
  Best FE silhouette index & 0.3530  \\
  Interpretability score 	  & 0.1952  \\
  Integrity score 		  & 0.1765  \\
  Chosen method 		  & SELECTION  \\  
  Number of selected features &  \\  
  to obtain target resolution & 7  \\
  Resolution 			  & 88.3\%  \\
  Best number of clusters (k) & 3  \\
\hline 
\end{tabular}
\end{minipage}\newline

The best FS silhouette index (0.3905) is greater than the value of the best FE silhouette index (0.3530). After applying the equations (\ref{eq:interpretability_score}) and  (\ref{eq:integrity_score}) using the parameters interpretability_oriented and integrity_oriented, the interpretability score (0.1952) is higher than the integrity score (0.1765). The selection process is used.

If the interpretability and the integrity are equally important, the nature of the data will determine which process does the best at generating a good SI (a good clustering consistency). This plays a role in equations (\ref{eq:interpretability_score}) and (\ref{eq:integrity_score}).

To reach 85\% of the resolution, we must use 7 features, having a resolution of 88.3\%. In the clustering process, the optimal number of clusters is 3. The SI figure and the stacked radar graphic are the same as in scenario 1 (Fig. \ref{fig:Scenario1_radar_multi_A} and Fig. \ref{fig:Scenario2_radar_multi_A}). \newline

\noindent \textbf{Case 4: Interpretability oriented and low resolution of data} \newline
This scenario is oriented toward interpretability. Compared to scenario 1, the resolution value has been lowered.  The value of interpretability_oriented = 0.9, integrity_oriented = 0.1 and target_resolution = 50\% has been used.  The results are shown in Table \ref{table:Scenario4}. \newline

\begin{minipage}[htb]{1.0\columnwidth}
\centering
\captionof{table}{Algorithm results using interpretability_oriented = 0.9, integrity_oriented = 0.1 and target_resolution = 50\%}%
\label{table:Scenario4}
\begin{tabular}{|l|r|}
\hline 
\textbf{Metrics} & \textbf{Values} \\ 
\hline 
  Best FS silhouette index & 0.4393  \\
  Best FE silhouette index & 0.3775  \\
  Interpretability score 	  & 0.3953  \\
  Integrity score 		  & 0.0377  \\
  Chosen method 		  & SELECTION  \\
  Number of selected features &  \\  
  to obtain target resolution & 4  \\
  Resolution 			  & 52.3\%  \\
  Best number of clusters (k) & 3  \\
\hline 
\end{tabular}
\end{minipage}\newline

The value of the best FS silhouette index (0.4393) is greater than the value of the best FE silhouette index (0.3775). Same as in scenario 1, the chosen method is the feature selection because the parameter interpretability_oriented (0.9) has a higher value than the integrity_oriented value (0.1) and the interpretability score (0.3953) is higher than the integrity score (0.0377). To reach 50\% of the resolution, we must use the best 4 features.  Those features have a resolution of 52.3\%. Doing the clustering process, the optimal number of clusters is 3. Fig. \ref{fig:Scenario4_Silhouette} shows the distribution of each element according to its silhouette index. 

\centering
\noindent \includegraphics[width=10cm]{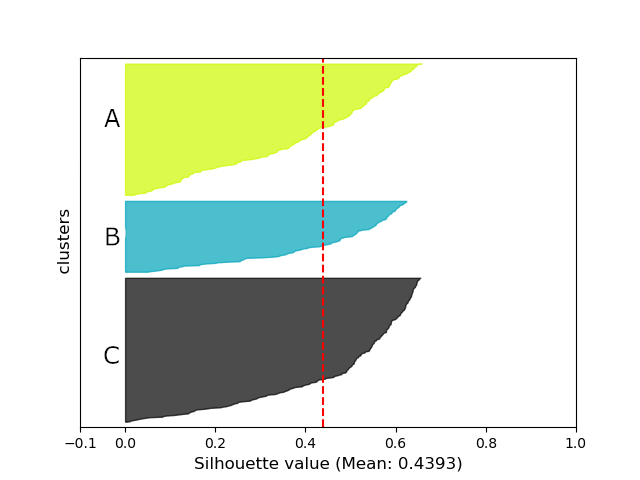}%
\captionof{figure}{Silhouette graphic for case 4 showing the consistency for clusters A, B, and C.\newline \label{fig:Scenario4_Silhouette}}
\justifying

This figure shows the 3 different clusters. This graphic shows no misplaced values (negative values). It also shows an average of 0.4393 (red dotted line).\newline

Fig. \ref{fig:Scenario4_radar_multi_A} shows the representation of the clustering using a stacked radar graphic.  

\centering
\noindent \includegraphics[width=10cm]{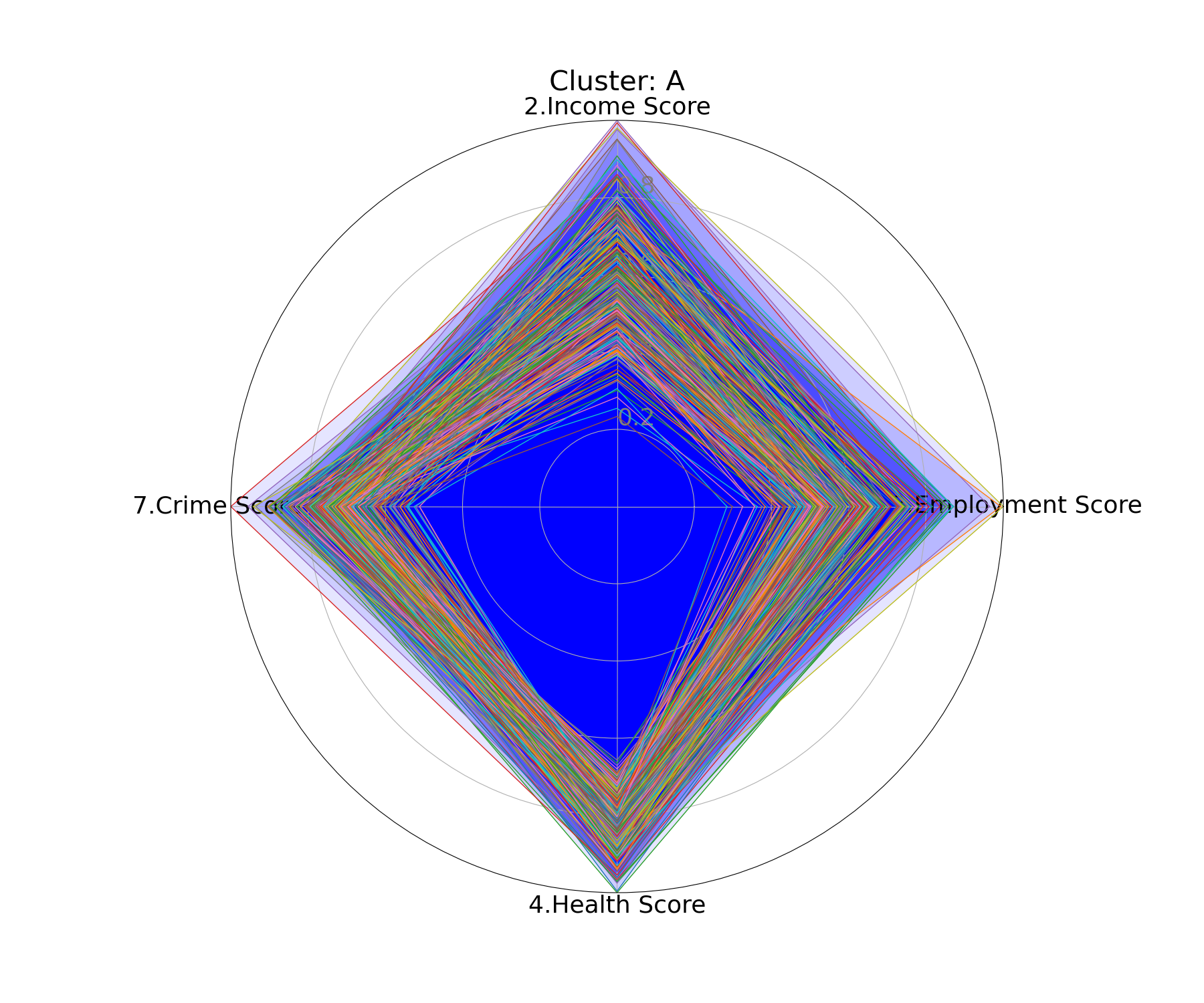}%
\captionof{figure}{Stacked radar graphics showing cluster A of normalized values for case 3\newline \label{fig:Scenario4_radar_multi_A}}
\justifying

The feature's names have been kept since this scenario is oriented toward interpretability. Compared to scenario 1 using a good resolution of data (7 features), this graphic shows only 4 features since the resolution value has been lowered to 50\%. At a glimpse, we can see that there is a good consistency. It is even a better consistency than in scenario 1 (SI = 0.4392 for scenario 4 and SI = 0.3905 for scenario 1). Remind that a better consistency is often linked to fewer dimensions in the data. \newline
 
\noindent \textbf{Case 5: Integrity oriented and low resolution of data} \newline
For this last scenario, integrity is more important than keeping the signification of the features, but a lower feature resolution than in scenario 2 is defined.  The values of interpretability_oriented = 0.1, integrity_oriented = 0.9 and target_resolution = 50\% has been set. Table \ref{table:Scenario5} shows the results using this configuration. \newline

\begin{minipage}[htb]{1.0\columnwidth}
\centering
\captionof{table}{Algorithm results using interpretability_oriented = 0.1, integrity_oriented = 0.9 and target_resolution = 50\%}%
\label{table:Scenario5}
\begin{tabular}{|l|r|}
\hline 
\textbf{Metrics} & \textbf{Values} \\ 
\hline 
  Best FS silhouette index & 0.4393  \\
  Best FE silhouette index & 0.3775  \\
  Interpretability score 	  & 0.0439  \\
  Integrity score 		  & 0.3397  \\
  Chosen method 		  & EXTRACTION  \\  
  Number of PCs 		  & 4  \\
  Resolution 			  & 54.2\%  \\
  Best number of clusters (k) & 3  \\
\hline 
\end{tabular}
\end{minipage}\newline

The best FS silhouette index (0.4393) is greater than the value of the best FE silhouette index (0.3775). The interpretability score is low (0.0439) and the integrity score (0.3397) is high. A feature extraction process is selected.  4 PCs are required to reach 50\% of the resolution.  The optimal number of clusters is 3. Fig. \ref{fig:Scenario5_Silhouette} shows the distribution of the SI for this case.

\centering
\noindent \includegraphics[width=10cm]{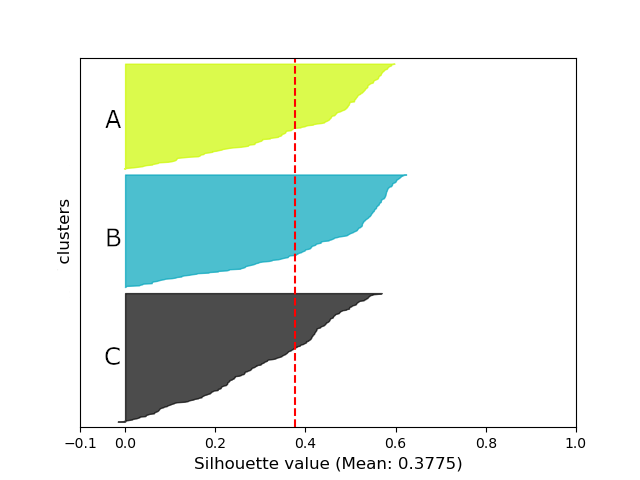}%
\captionof{figure}{Silhouette graphic for case 5 showing the consistency for clusters A, B, and C.\newline \label{fig:Scenario5_Silhouette}}
\justifying

This figure shows 3 clusters without only one misplaced value (between -1 and 0). The average of the SI is 0.3775, which shows a significantly better SI than in case 2 (0.353), which has more dimensions. \newline

Fig. \ref{fig:Scenario5_radar_multi_A} presents a stacked radar graphic of  cluster A. 

\centering
\noindent \includegraphics[width=10cm]{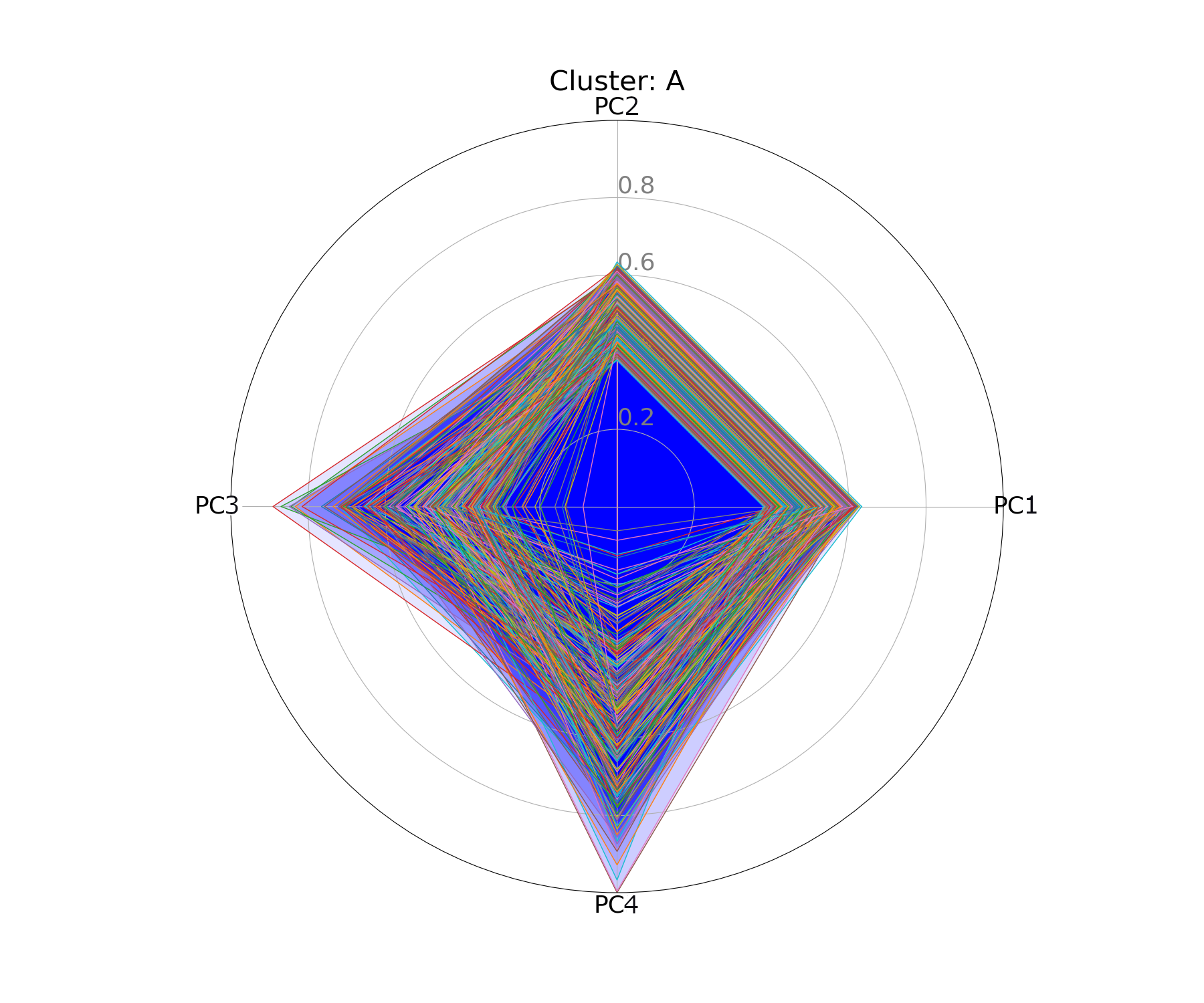}%
\captionof{figure}{Stacked radar graphics showing cluster A of normalized values for case 5\newline \label{fig:Scenario5_radar_multi_A}}
\justifying

The consistency is quite good. Although, like all the clusters whose features got throw a feature extraction process, the feature's name is lost and replaced by PCs. The interpretability is then reduced.  

\subsection{Method validation}%
\label{subsec:SubResultsValidation}%
This last part of the results is the validation of the method. To ensure that the algorithm takes the right decision, 250 realistic random cases have been generated. Each of the random cases includes a random SI index (after a hypothetical feature selection), a random SI index (after a hypothetical feature extraction), and a random interpretability importance parameter. An integrity importance parameter has also been computed using 1 - (the interpretability importance) parameter. Using those data, the decision algorithm has been called. For each case, an interpretability score and an integrity score have been calculated.  A decision has been finally taken between a feature selection or a feature extraction process.  Fig. \ref{fig:Validation1} shows the classification of the points according to the interpretability scores and the integrity scores.  The red points will use a feature extraction process and the blue points will use a feature selection process. The black line divides the interpretability (feature selection) and the integrity (feature extraction) domains. 

\centering
\noindent \includegraphics[width=10cm]{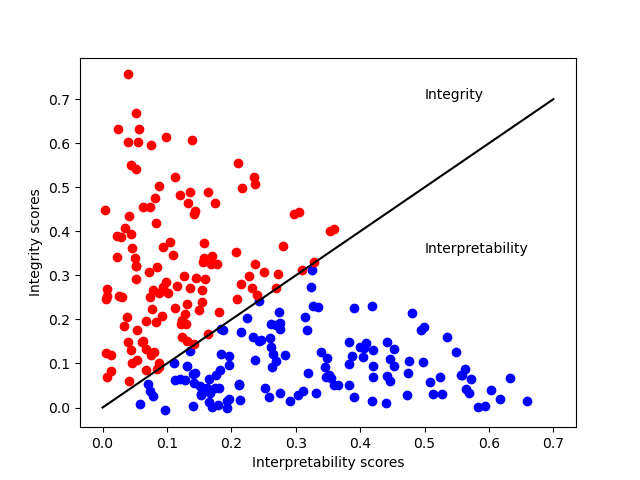}%
\captionof{figure}{Decision distribution classified in two groups: Interpretability and integrity\newline \label{fig:Validation1}}
\justifying

The points can't have a high value on both axes. This is because of the interpretability importance parameters that must be the inverse of the integrity importance parameter (\textit{alpha} and \textit{1 - alpha}). Those are used in \ref{eq:interpretability_score} and \ref{eq:integrity_score}, which are the axis. If one value is very high, the other must be very low. Both can have an average value. This graphic shows that the algorithm takes always a good decision even when it would not be easy for a human to choose. Since the algorithm uses a threshold, the classification is always correct. This graphic is simple, but it validates the results of the complex previous parts of the process that uses FRSD and PCA. \newline

Fig. \ref{fig:Validation2} shows the bar pairs of the number of features (blue) and the principal components (red), according to the target resolution of data (as specified in the parameters). As in the use cases, the London dataset has been used. 

\centering
\noindent \includegraphics[width=10cm]{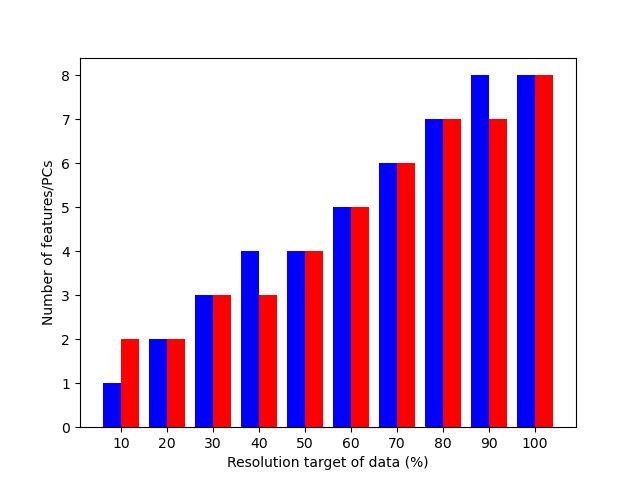}%
\captionof{figure}{Number of features/PCs according to the target resolution of data.\newline \label{fig:Validation2}}
\justifying

\noindent As expected, we can see that a perfect resolution of 100\% requires all of the 8 available features. This number slowly declines when subtracting each step of 10\%.  To validate the integrity advantage of the feature extraction over the feature selection, we must subtract their respective resolutions. It can be compared only when they have the same number of features and principal components.  Fig. \ref{fig:Validation3} displays the difference percentage  for all the target resolutions having the same amount of features/PCs. For instance, reading the Fig. \ref{fig:Validation2} we can see that resolutions of 20\%, 30\%, 50\%, 60\%, 70\%, 80\% and 100\% have the same number of features/PCs. This is where the values of Fig. \ref{fig:Validation3} are defined. Blue bars represent the resolution percentage differences between feature extraction and feature selection.

\centering
\noindent \includegraphics[width=10cm]{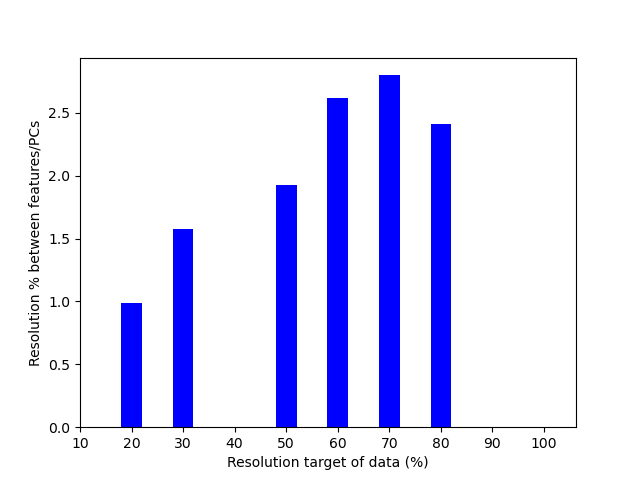}%
\captionof{figure}{Differences of resolution in percentage between feature extraction and feature selection.\newline \label{fig:Validation3}}
\justifying

We can see that there is a resolution advantage when using feature extraction. This validates the integrity-oriented parameter. 

As for the interpretability-oriented parameter, the best way to validate it is simply by comparing any graphs after a feature selection and a feature extraction.  For instance, let's compare Fig. \ref{fig:Scenario1_radar_multi_A} to Fig. \ref{fig:Scenario2_radar_multi_A}. It is easier to interpret real features names as in Fig. \ref{fig:Scenario1_radar_multi_A} than it is to interpret abstracts principal components (PC1, PC2...) as in Fig. \ref{fig:Scenario2_radar_multi_A}.  This validates the interpretability-oriented parameter. 

\section{Discussions}%
\label{sec:Discussions}%
Choosing to reduce the dimensionality or not has always been a critical decision. Using the right technique to reduce the dimensionality of a set of features is also an important decision. Especially in an unsupervised learning context, where data labels are not available.  The main contribution of this paper is to define a new complete method that makes the right decision of dimensionality reduction according to the data scientist preferences, before completing the process by clustering the data. 

Two different algorithms have been used to evaluate the feature's importance: FRSD and PCA. The first evaluates the feature importance for feature selection and the second for feature extraction. In Tables \ref{table:Feat_importance_FRSD} and \ref{table:Feat_importance_PCA}, it is notable that both methods give similar results in terms of features importance. The principal difference is the loss of the feature's name when using PCA. 

The decision process uses two equations, (\ref{eq:interpretability_score}) and (\ref{eq:integrity_score}).  Both are based on the data scientist's preferences regarding interpretability and integrity. The decision equations are also based on the best SI (a metric of cluster consistency), and the previously calculated feature's importance. Comparing the best SI for feature selection and feature extraction in Table \ref{table:Scenario1}, we note that feature extraction does not mean a better consistency of clustering. Reading Table \ref{table:Scenario3} where the values of the parameters  (oriented_interpretability and oriented_integrity) are equal, we can see that using a method that keeps better integrity of data (like PCA) does not guarantee a better consistency of clustering.  Better consistency of clustering is given by scenarios 4 and 5. That is because of the lower data resolution.  Both feature selection and feature extraction allow a better cluster consistency when the dimension is reduced. A higher resolution of data (more features) implies that it is harder to keep a good cluster consistency.  

PCA is a useful tool to extract features and reduce dataset dimensionality. Consequentially, it helps to speed up the learning process and to simplify the presentation of the features. Before extracting some features, it is very important to evaluate the impact of such an operation.  In some cases, features can be extracted without losing too much precision in the data. Although, in some other cases, significant resolution of the data will be lost.  The feature extraction has the disadvantage of losing the name of the features. Consequently, a clustering process after a feature extraction can give good results, but it becomes less significant when represented on figures as in Fig. \ref{fig:Scenario2_radar_multi_A}, since the feature's names are lost. Having axis named PC1, PC2, PC3... PCn is harder to interpret. 
Using a feature selection allows to keep the feature's names as in Fig. \ref{fig:Scenario1_radar_multi_A}, but at the price of losing some information. The trade-off must be carefully evaluated. This trade-off is precisely evaluated using this proposed method. 
 
The final result is clustering. For each scenario from 1 to 5, 2 types of graphics are presented to represent the clustering process. 1. The SI figures (Fig. \ref{fig:Scenario1_Silhouette}, \ref{fig:Scenario2_Silhouette}, \ref{fig:Scenario4_Silhouette}, and \ref{fig:Scenario5_Silhouette}) show the distribution of the data in each cluster. 
It shows the SI average, the number of clusters, the number of elements in each cluster, and the consistency of each cluster. 2. The stacked radar graphics (Fig. \ref{fig:Scenario1_radar_multi_A}, \ref{fig:Scenario2_radar_multi_A}, \ref{fig:Scenario4_radar_multi_A}, and \ref{fig:Scenario5_radar_multi_A}) that displays the normalized values of each feature or PC. Reading those graphics, it is possible to note de cluster consistency at a glimpse. Only one cluster per scenario is displayed (cluster A), as an example.

The validation of the method is shown in Fig. \ref{fig:Validation1}, \ref{fig:Validation2}, and \ref{fig:Validation3}.  Fig. \ref{fig:Validation1} shows that the algorithm takes the good decision of feature selection or feature extraction using a set of 250 generated data and parameters. Fig. \ref{fig:Validation2} shows a good link between the target resolution parameters and the number of selected features and PCs. Finally, Fig. \ref{fig:Validation3} shows the advantage of using feature extraction, in terms of the integrity of data. 

This research is a complement to the recent FRSD method presented in \cite{yu_ensemble_2020}. This methodology proposed a new method to evaluate the features in an unsupervised learning clustering context. Based on this work, we can compare the added value of the present paper.  This paper takes this useful method and includes it in a more global and integrated context where FRSD and PCA are used to evaluate the importance of the features. From this evaluation and according to some parameters on interpretability and on integrity of features, a score is calculated and used to decide whether a feature selection or a feature extraction is the best in this situation.  In most cases, the utility of calculating the importance of the features with FRSD is to reduce dimensionality and to apply a clustering process.  The reason is that FRSD aims to determine the importance of the feature relative to the consistency of the clustering process (the SI). FRSD and a clustering algorithm like K-Means are linked. This method adds to FRSD the whole chain process, from the evaluation of the features to the final clustering and the representation of the data (SI and stacked radar graphs). 

\section{Conclusion}%
\label{sec:Conclusion}%

This paper proposed a new method to evaluate and select the right technique of dimensionality reduction and to evaluate the importance of each feature regarding the consistency of the clustering process. It then applies this selected technique and proceeds to the clustering and its representations.  \newline
In the future, a similar method could be developed to automatize de choice of  dimensionality reduction, but in a supervised learning context.  It would be a very similar method. A Random Forest algorithm would replace the FRSD algorithm since the label of the data is already known in a supervised learning context. There would be no need to apply clustering afterward since this would be in a supervised learning process. Also, this method could be validated using more data from different sizes. Different presentations of the data can also be developed.   

\section{Acknowledgement}%
\label{sec:Acknowledgement}%
This work has been supported by the "Cellule d’expertise en robotique et intelligence artificielle" of the Cégep de Trois{-}Rivières.

%
\bibliographystyle{plain}%
\bibliography{paper}

%
\end{document}